\documentclass[a4paper,conference]{IEEEtran}
\IEEEoverridecommandlockouts

\ifCLASSINFOpdf
  \usepackage[pdftex]{graphicx}
  \DeclareGraphicsExtensions{.pdf,.jpeg,.png}
\else
  \usepackage[dvips]{graphicx}
  \DeclareGraphicsExtensions{.eps}
\fi

\usepackage{xcolor}
\usepackage{amsmath}

\usepackage{amsfonts}
\usepackage{amsthm}
\usepackage{algorithm}
\usepackage{algpseudocode}
\usepackage{array}
\usepackage{url}
\usepackage{amsfonts}

\newtheorem{Definition}{Definition}
\newtheorem{proposition}{Proposition}

\begin{document}

\title{Hcore-Init: Neural Network Initialization based on Graph Degeneracy}

\author{\IEEEauthorblockN{
\color{red}{TODO:na valoume onomata(single blind review)}
\color{black}{}
Anonymous Authors}}

\author{

\IEEEauthorblockN{Stratis Limnios$^1$\thanks{$^1$ Supported by ANR project ESIGMA (ANR-17-CE23-0010).}}
\IEEEauthorblockA{\textit{\'Ecole Polytechnique}\\
Palaiseau, France}

\and
\IEEEauthorblockN{George Dasoulas}
\IEEEauthorblockA{\textit{\'Ecole Polytechnique} $\&$\\
\textit{Noah's Ark Lab, Huawei Technologies}\\
Paris, France}

\and
\IEEEauthorblockN{Dimitrios M. Thilikos$^{1,2}$\thanks{$^2$ Supported by ANR project DEMOGRAPH (ANR-16-CE40-0028).}}
\IEEEauthorblockA{\textit{LIRMM, Univ Montpellier, CNRS}\\
Montpellier, France}

\and
\IEEEauthorblockN{Michalis Vazirgiannis$^1$}
\IEEEauthorblockA{\textit{\'Ecole Polytechnique}\\Palaiseau, France}

}

\maketitle

\begin{abstract}
Neural networks have become a very popular tool for many machine learning tasks, as in recent years we witnessed many novel architectures, learning and optimization techniques for deep learning.  Capitalizing on the fact that neural networks inherently constitute multipartite graphs among neuron layers, we aim to analyze directly their structure to extract meaningful information that can improve the learning process. 
To our knowledge graph mining techniques for enhancing learning in neural networks have not  been thoroughly investigated. In this paper we propose an adapted version of the k-core structure for the complete weighted multipartite graph extracted from a deep learning architecture. 
As a multipartite graph is a combination of bipartite graphs, that are in turn the incidence graphs of hypergraphs, we design k-hypercore decomposition, the hypergraph analogue of  k-core degeneracy. 
We applied k-hypercore to several neural network architectures, more specifically to convolutional neural networks and multilayer perceptrons for image recognition tasks after a very short pretraining. Then we used the information provided by the hypercore numbers of the neurons to re-initialize the weights of the neural network, thus biasing the gradient optimization scheme. Extensive experiments proved that k-hypercore outperforms the state-of-the-art initialization methods. 

\end{abstract}


%
\IEEEpeerreviewmaketitle

\section{Introduction}

During the last decade deep learning has been intensely in the focus of the research community. Its applications on a huge variety of scientific and industrial fields highlighted the need for new approaches at the level of neural network design. 
Researchers have studied until today different aspects of the Neural Network (NN) architectures and how these can be optimal for various tasks, i.e the optimization method used for the error backpropagation, the contribution of the activation functions between the NN layers or normalization techniques that encourage the loss convergence, i.e batch normalization, dropout layer etc. 

Weight initialization is one of the aspects of NN model design that contribute the most to the gradient flow of the hidden layer weights and by extension to the ability of the neural network to learn. The main focus on the matter of weight initialization (\cite{Glorot10understandingthe}, \cite{he2015delving}) is the observation that weights among different layers can have a high variance, making the gradients more likely to explode or vanish. 
\begin{figure}
    \centering
    \includegraphics[scale = 0.3]{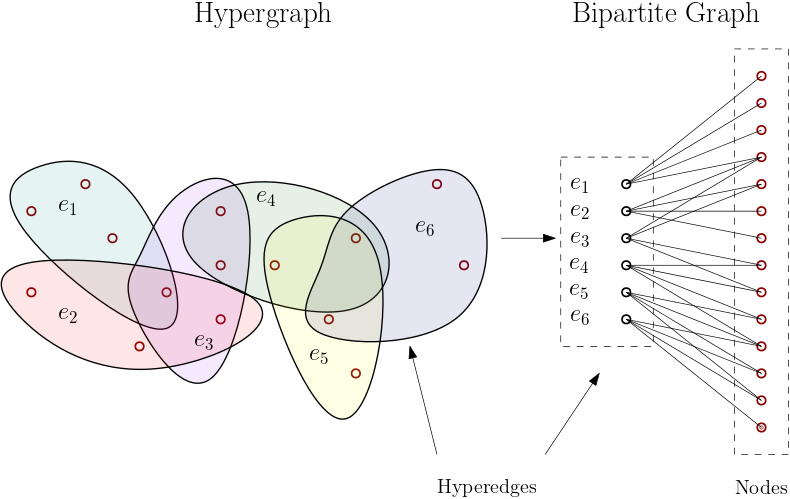}
    \caption{Hypergraph and the corresponding incidence graph}
    \label{fig:hyptobip}
\end{figure}
Neural Networks capitalize on graph structure by design. Surprisingly there has been very little work analyzing them as a graph with edge and/or nodes attributes. Recent work \cite{skianis2019rep}   introduces graph metrics to produce latent representation sets capitalising on bipartite matching directly implemented in the neural network architecture, which proved to be a very powerful method. Also the work by C. Morris analyzes the expressivity of Graph Neural Networks using the Weisfeiler-Leman isomorphism test \cite{morris2019weisfeiler}. Our interest lies in trying to refine the optimization scheme by capitalizing on graph metrics and decompositions. One natural candidate was the $k$-core decomposition \cite{seidman1983network}. Indeed this decomposition method, being very efficient ($O(n \log(n))$ in the best cases \cite{batagelj2003m}), performs very well in state-of-the-art frameworks for enhancing supervised learning methods \cite{nikolentzos2018degeneracy}. Providing key subgraphs, and also extracting very good features. 

\begin{figure}
    \centering
    \includegraphics[scale = 0.25]{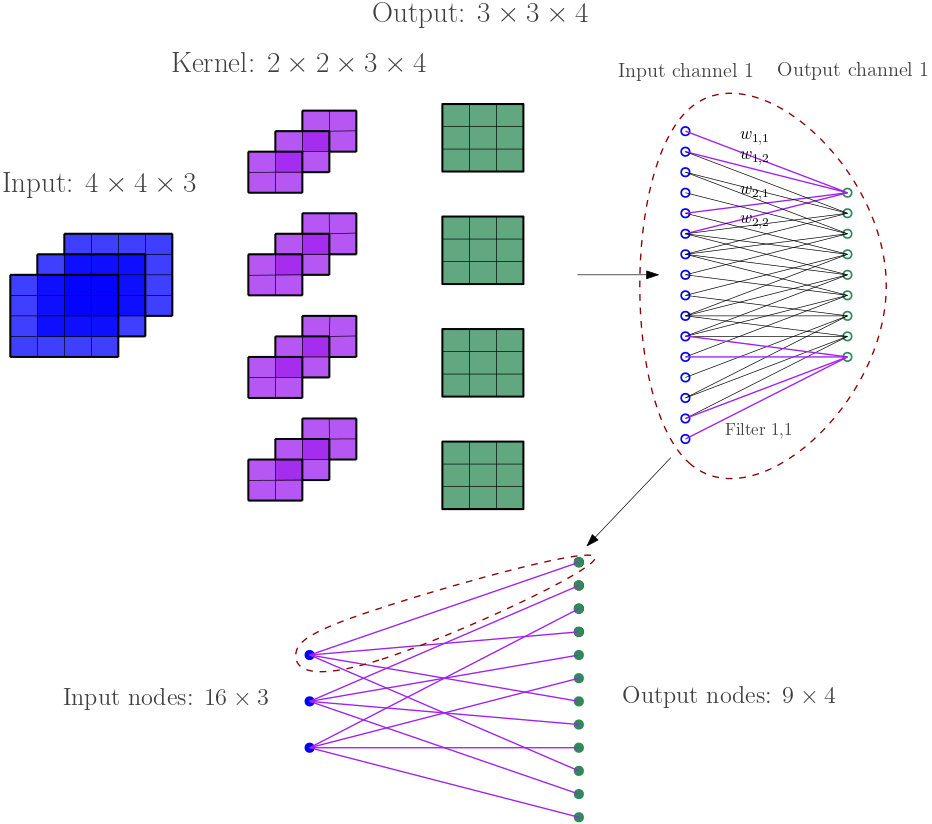}
    \caption{Illustration of the transformation of a CNN to graph.}
    \label{fig:cnntograph}
\end{figure}

Unfortunately, in the case of a graph representing a neural network, $k$-core might lack some features. As a matter of fact, graphs extracted from NNs constitute multipartite complete weighted graphs, in the case of a Multilayer Perceptron and almost complete for Convolutional Neural Networks. As we saw different $k$-core variants for different types of graphs, such as the $k$-truss \cite{rossi2015spread} counting triangles, the $D$-core \cite{giatsidis2013d} for directed graphs, were designed this past decade. A natural thought was then to design our own version of the $k$-core for our precise graph structure. \\
Hence our contributions are the following:
 
\begin{itemize}
    \item We provide a unified method of constructing the graph representation of a neural network as a block composition of the given architecture (see fig. \ref{fig:cnntograph}). This is achieved by transforming each part of the network (i.e linear or convolutional layers, normalization/dropout layers and pooling operators) into a subgraph. Having this graph representation, it is possible to apply different types of combinatorial algorithms to extract information from the graph structure of the network.       
    \item Next we design a new degeneracy framework, namely the \textbf{$k$-hypercore}, extending the concept of k-core to bipartite graphs by considering that each pair of layers of the neural network, constituting a bipartite graph, is the incidence graph of a hypergraph (see fig.\ref{fig:hyptobip}).
    \item we propose a novel weight initialization scheme, \textbf{Hcore-init} by using the information provided by the weighted version of the \textbf{$k$-hypercore} of a NN extracted graph, to re-initialize the weights of the given neural network, in our case, a Convolutional neural network and a Multilayer Perceptron. Our proposal clearly outperforms traditional initialization methods on classical deep learning tasks.
\end{itemize}
The rest of this paper is organized as follows, first some preliminary definitions and overview of the state of the art methods in neural network initialization methods. Then we provide the methodology which allows us to transform neural networks to edge weighted graphs. Further on, we proceed to the main contribution of the paper being the definition of the hypercore degeneracy and the procedure which produces our initialization method. Finally we test our method on several image classification datasets, comparing it the main initialization method used in neural networks. 


\section{Preliminaries}
In deep neural networks, weight initialization is a vital factor of the performance of different architectures \cite{mishkin2015all}. The reason is that an appropriate initialization of the weights of the neural network can avert the explosion or vanishing of the layer activation output values.
\subsection{Initialization methods}
\subsubsection{Glorot Initialization}
One of the most popular initialization methods is Glorot initialization \cite{Glorot10understandingthe}. According to that, the weights of the network are initialized by drawing them from a normal distribution with $E[W] = 0$ and ${\sf Var}(w_i) =\frac{1}{\mathrm{fanin}}$, where $\mathrm{fanin}$ is the number of incoming neurons. Also, more generally, we can define variance with respect to the number of outgoing neurons as: ${\sf Var}(w_i) =\frac{1}{\mathrm{fanin}+\mathrm{fanout}}$, where $\mathrm{fanout}$ is the number of neurons that the output is directed to.\\

\subsubsection{Kaiming He Initialization}
Although Glorot initialization method manages to maintain the variance of all layers equal, it assumes that the activation function is linear. In most of the cases of non-linear activation function that Glorot initialization is used, the hyperbolic tangent activation is employed. The need for taking into account the activation function for the weight initialization led to the Kaiming He Initialization \cite{he2015delving}. According to this method, in the case that we employ \textit{ReLU} activation functions,  we initialize the network weights by drawing samples from a normal distribution with zero mean: $E[W] = 0$ and variance that depends on the order of the layer: ${\sf Var}[W] = \frac{2}{n^l}$, where $l$ is the index of the $l$-th layer and $n$ the number of neurons in the given layer.

One main assumption for weight initialization is that the mean of the random distribution used for initialization needs to be $0$. Otherwise, the calculation of the variances presented above could not be done and we won't be able to have a fixed way to initialize the variance. \\
Since in our work we want to capitalize on the $k$-hypercore decomposition to \textit{bias} those distributions we will have to face the fact that we might not be able to control the variance of the weights we initialize. Thankfully the fact that the initial distribution has $0$ mean will ensure that our method respects as well this condition on every layer of the neural network.\\

Moreover, since the $k$-hypercore decomposition is defined over hypergraphs, let us recall some properties of hypergraphs and their relations with bipartite graphs.

\subsection{Hypergraphs and Bipartite graphs}

A hypergraph is a generalization of a graph in which an edge can join any number of vertices. It can be represented and we keep this notation for the rest of the paper as $\mathcal{H}=(V,E_{\cal H})$ where $V$ is the set of nodes, and $E_{\cal H}$ is the set of hyperedges, i.e. a set of subsets of $V$. Therefore $E_{\cal H}$ is a subset of $\mathcal{P}(V)$. Moreover a bipartite graph is the incidence graph of a hypergraph \cite{pisanski2000bridges}. Indeed, a hypergraph $\mathcal{H}$ may be represented by a bipartite graph $\mathcal{G}$ as follows: the sets $X$ and $E$ are the partitions of $\mathcal{G}$, and $(x_1, e_1)$ are connected with an edge if and only if vertex $x_1$ is contained in edge $e_1$ in $\mathcal{H}$. Conversely, any bipartite graph with fixed parts and no unconnected nodes in the second part represents some hypergraph in the manner described above. Hence, we can consider that every pair of layers in the neural network can be viewed as a hypergraph, where the left layer represents the hyperedges and the right the nodes (see fig.\ref{fig:hyptobip}).

\section{Graph Extraction from Neural Network Architecture}


We will now describe how we map the two classic neural network architectures we investigate to graphs, and more specifically to a collection of bipartite ones. Also, from now on, we are going to refer to a fully-connected neural network as FCNN and to a convolutional neural network as CNN \cite{krizhevsky2012imagenet}.

\subsection{Fully-Connected Neural Networks}

Let a FCNN $\mathcal{F}$ with $L$ hidden layers, $n_i$,  $i=1,..,L$ number of hidden units per layer and $W_i \in \mathbb{R}^{n_i,n_{i+1}}$ the weight matrix of the links between the units of the layers $i$ and $i+1$.\\
We define the graph $G_{\mathcal{F}}=(V, E, W)$ as the graph representation of the FCNN $\mathcal{F}$, where the set of nodes $V$ corresponds to the $\sum_{i=1}^L n_i$ number of hidden units of $\mathcal{F}$, the set of edges $E$ contains all the links of unit pairs across the layers of $\mathcal{F}$ and the edge weight matrix $W$ corresponds to the link weight matrices $W_i, i=1,..., L-1$. We note that the graph representation $G_{\mathcal{F}}$ does not take into account any activation functions $\sigma$ used in $\mathcal{F}$.\\
\textbf{Remark.} It is easy to see that $G_{F}$ is a $k$-partite graph (i.e a graph whose vertices can be partitioned into $k$ independent sets) and more specifically a union of $L-1$ complete bipartite graphs.

\subsection{Convolutional Neural Networks}
After showing the correspondence between a FCNN $\mathcal{F}$ and its graph representation $G_\mathcal{F}$, we are ready to define the graph representation of a CNN layer. Let a CNN layer $\mathcal{C}$. The convolutional layer is characterized by the input information that has $I$ input channels where each channel provides $n\times n$ features (i.e an $24\times 24$ image characterized by the 3 RGB channels), the output information that has $O$ output channels, where each channel has $m\times m$ features and the matrix of the convolutional kernel $F \in \mathbb{R}^{w \times h \times I \times O}$, where $w,h$ are the width and height of the kernel.\\
In order to define the graph $G_\mathcal{C} = (V,E,W)$ as the graph representation of the CNN $\mathcal{C}$, we have to flatten the 3 and 4-dimensional input, output, and filter matrices correspondingly. Specifically, the $G_\mathcal{C}$ is a bipartite graph, where the first partition of nodes $P_1$ is the flattened input information of the CNN layer  ( $|P_1| = I \times n \times n$ ) , the second partition of nodes $P_2$ is the flatten output information ($|P_2| = O \times m \times m$).

\section{Weight initialization based on Hcore}

As degeneracy frameworks have proven to be very efficient at extracting influential individuals in a network \cite{al2017identification}, we are motivated to consider  structural information provided by the hcore decomposition of the network to identify ``influential" neurons.

Assuming a neural network graph, we provide a definition of degeneracy specifically for hypergraphs, where standard $k$-core does not apply.

\begin{Definition}[Hypercore]
Given a hypergraph $\mathcal{H}=(V,E_{\cal H})$ We define the $(k,l)$-\textbf{hypercore} as a maximal connected subgraph of $\mathcal{H}$ in which all vertices have hyperdegree at least $k$ and all hyperedges have at least $l$ incident nodes.
\end{Definition}

As for now on, we will refer to the $(k,2)$-hypercore as the $k$-hcore and similarly, the \textit{hcore number} of the node will be the largest value of $k$ for which the given node belongs to the $k$-hcore.

\begin{figure}
    \centering
    \includegraphics[scale = 0.20]{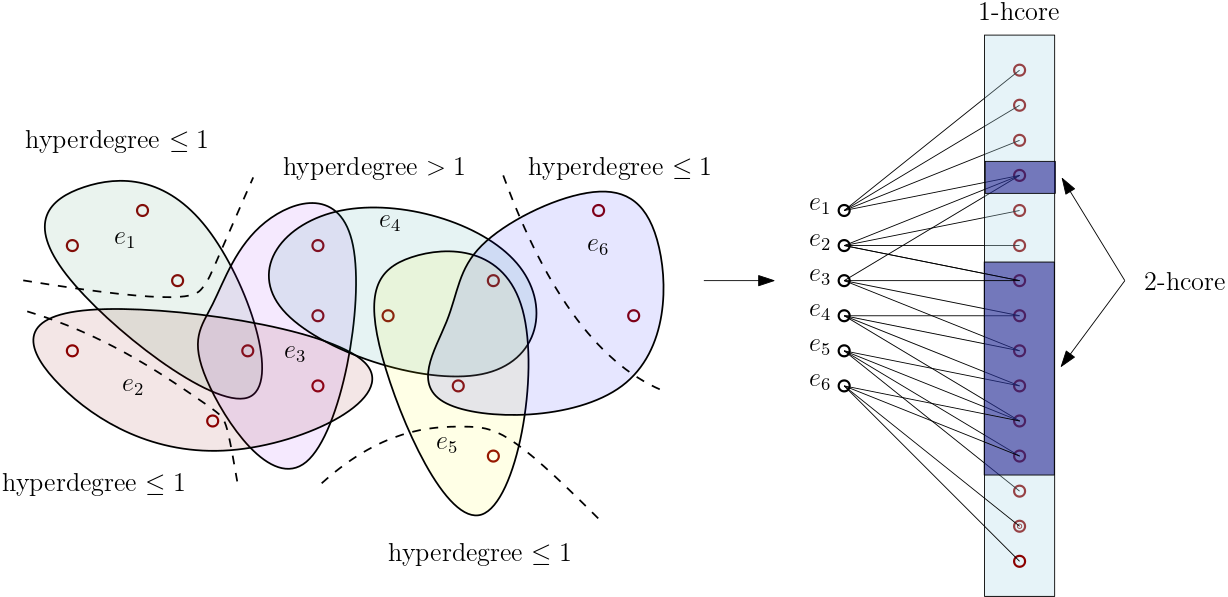}
    \caption{Example of a $k$-hcore decomposition of a hypergraph}
    \label{fig:hcore_ex}
\end{figure}

This provides a hypergraph decomposition and in our case a decomposition of the right handside of the studied bipartite graph (see fig. \ref{fig:hcore_ex}), as we do not care about the hcore of the hyperedges. \\

\begin{algorithm} 
 \caption{Hcore decomposition algorithm}{\label{hcore}}
\begin{algorithmic}[1]
\Procedure{Hcore}{$G,rnodes$}
\State{\textbf{Input} $G$: bipartite graph, $rnodes$: right layer nodes}
\State{\textbf{Output} $hcore$: dictionary of hcore values}
\State
\State $hcore$ $\gets$ $dict$($(node,0)$ for $node$ in $rnodes$)
\State $tokeep \gets rnodes$
\While{$tokeep \neq \emptyset$}
    \State $state \gets True$
    \While{$state == True$}
        \State $state \gets False$
        \State $tokeep \gets []$
        \For{$node \in rnodes$}
            \If{$G.degree(node)>k$}
                \State $tokeep.append(node)$
            \Else
                \State    $hcore[node]=k$
                \State    $graph.remove[node]$
                \State    $state \gets True$
            \EndIf
        \EndFor                
        \For{$node \in G.nodes\setminus rnodes$}
            \If{$G.degree(node)=1$}
                \State $G.remove(node)$
            \EndIf
        \EndFor
    \EndWhile
    \State $k \gets k+1$
\EndWhile
\EndProcedure

\end{algorithmic}
\end{algorithm}


Since we deal with edge-weighted bipartite graphs, we will use the weighted degree to define the hcore ranking of the nodes given the following weighted-hypercore definition:

\begin{Definition}[Weighted-hypercore]
Given an edge weighted hypergraph $\mathcal{H}=(V,E_{\cal H})$, we define the $(k,l)$-\textbf{weighted-hypercore} as a maximal connected subgraph of $\mathcal{H}$ in which all vertices have hyper-weighted-degree at least $k$ and all hyperedges have at least $l$ incident nodes.
\end{Definition}
Again, we will refer to the $(k,2)$-weighted-hypercore as $k$-WHcore. 
Now that we have this weighted version, we need to define a way to initialize the weights of the neural network. Indeed, since the WHcore is a value given to the nodes of the network and not the edges, being the weights we aim to initialize.
The WHcore shows us which neurons gather the more information, positive on the one hand and negative on the other. After a quick pretraining, we learn the weights just enough to show which neurons have a higher impact on the learning. This information is then grouped by the WHcore into influential neurons and less influential ones.\\
Moreover, since weights in neural networks are sampled from centered normal law, we have positive and negative weights. Since the WHcore framework operates on positive weighted degrees, we provide two graph representations of the neural network, namely $G^+$ and $G^-$. The $G^+$ graph is built upon the positive weights of the neural network, and the edge weights of the $G^-$ graph are the absolute values of the negative weights of the neural network. Indeed if between neuron $x_i$ and neuron $y_j$, $w_{ij}>0$  then we add an edge with weight $w_{ij}$ between node $x_i$ and $y_j$ in graph $G^+$, otherwise we add an edge with weight $|w_{ij}|$ between node $x_i$ and $y_j$ to graph $G^-$.\\
\textbf{Remark.} It is important to note that the \textbf{WHcore number} of a node is the largest $k$ in which a node is contained in the $k$-WHcore. 
Also, the WHcore number of a node is a function of the degree of the node. As the degree depends on the weights, there exists two functions $g$ and $h$ such that $g(W,x)$ outputs the weighted degree of a node x, thus being a linear combination of the weights $W$. Then $c(W,x) = h(g(W,x))$ is the WHcore number of the node $x$. For convenience, we now write $c(W^+,x_k) = c^+_k$ where $W_k$ are the positive weights of the weight matrix $W$. \\
Moreover, the following initialization schemes are done after a small amount of pretraining of the neural network, in order to have preliminary information of the importance of the impact of the neurons.
\subsection{Initialization of the FCNN}

The initialization then is then dependent on the architecture we are looking at, indeed for an FCNN as the graph construction is fairly straightforward we proceed as follows:\\
For every pair of layers for both positive and negative graphs, we have nodes $x_i$, with $i\in \{1,\dots,\text{fanin}\}$ in the left side of the bipartite graph, and $y_j$ nodes, with $j\in\{1,\dots,\text{fanout}\}$ nodes on the right side. As for every node $y_i$ we compute their WHcore from the graph $G^-$, $c^-_j$ and from the graph $G^+$, $c^+_j$.
Then the given layer weights $w_{i,j}$, are initialized, depending on their sign, with a normal law with expectancy:

\begin{itemize}
    \item for all $i$ if $w_{i,j}\geq 0$, $M = \frac{c^+_j}{\sum_{1\leq k \leq \text{fanout}}c^+_k}$,
    \item else $M = \frac{c^-_j}{\sum_{1\leq k \leq \text{fanout}}c^-_k}$
\end{itemize}
and with the same variance used in Kaiming He initialization. We prove later that the overall mean value of the new random variable obtained in this fashion is $0$ as well, justifying the use of the Kaiming He variance to be optimal.

\subsection{Initialization of the CNN}

For the CNN, since the induced graph is more intricate and the filter weights must follow the same distribution, the initialization framework has to be adapted. We still compute the WHcore on a pair of layers but keeping the filters in mind, the left layer nodes are $x_i^{(k)}$ with $i\in \{1,\dots,n \times n\}$ the input size and $k\in \{1,\dots,I\}$ the number of input filters. Similarly the left layer nodes are $y_j^{(k')}$, where $j \in \{1, \dots, m\times m\}$ the output size, and $k'\in \{1,\dots,O \}$ the output channels. We remind as well that we have two WHcores, one for the positive graph $c^+$ and one for then negative $c^-$. Then for a given filter $w^{(k,k')}$ its values are initialized with the following method: 

\begin{itemize}
    \item we define $f$ for a given filter $W$ as $m(W^+) = \frac{1}{H^2}\sum_{j}c_j^+$ and $m(W^-) = \frac{1}{H^2}\sum_{j}c_j^-$,
    if $m(W^+)-m(W^+) > 0$ then $M = m(W^+)$
    \item else $M = -m(W^-)$.
\end{itemize}

Using the notations given in the previous remark we can write $m$ in the following general form : 
$$ m = \text{sign}(\text{argmax}(m(W^+),f(W^-))) \max(m(W^+),m(W^-))$$
where $\text{sign}(W^+) = 1$ and $\text{sign}(W^-) = -1$.

This initialization is done for every filter and with variance given by the Kaiming He initialization method. Now we will prove that for the CNN the overall expectancy of the mean value produced is indeed 0.

\begin{proposition}

Let $X_1$ and $X_2$ two centered i.i.d. random variables with symmetric distribution. We define $X^+ = \text{max}\{X_1,0\}$, $X^-=\text{max}\{X_2,0\}$, and a real valued measurable function $f:\mathbb{R}_+\rightarrow \mathbb{R}$ such that $\mathbb{E} [ | f(X^+) | ] <\infty$ and $\mathbb{E} [ | f(X^-) | ] <\infty$.\\

Then:
\begin{itemize}
    \item $X^+$, $\!X^-$ are positive i.i.d. random variables.
    \item The random variable:
    \begin{equation*}
Z = \text{sign} \big( \text{argmax}(f(X^+),f(X^-)) \big) \text{max}\big(f(X^+),f(X^-)\big)
\end{equation*}
is centered, i.e. $\mathbb{E}[Z]=0$.
\end{itemize}
\end{proposition}

\begin{proof}
We remind that the function $\mathbb{I}_{\{x\in X\}}$ is the Euler indicator function:

$$\mathbb{I}_{\{ x\in X\}} = \left\{
  \begin{array}{rcr}
    1  &  \hspace{1pt}  \hspace{1pt} \text{if~} x\in X\\
    0  &  \hspace{1pt} \text{\,\, otherwise.}\\
  \end{array}
\right.
$$

Let us proceed to evaluate the expectancy of $Z$ provided that $X^+$ and $X^-$ are i.i.d.:
\begin{eqnarray*}
     &&\!\!\!\!\!\!\!\!\mathbb{E}[Z] =\mathbb{E}[Z \mathbb{I}_{\{f(X^+) > f(X^-) \}}]  + \mathbb{E}[Z \mathbb{I}_{\{f(X^+) \leq f(X^-) \}}] =  \\
     &&\!\!\!\!\!\!\!\!\mathbb{E}[f(X^+)\mathbb{I}_{\{f(X^+) > f(X^-) \}}]- \mathbb{E}[f(X^-)\mathbb{I}_{\{f(X^+) \leq f(X^-) \}}]=\\
     &&\!\!\!\!\!\!\!\!  \mathbb{E}[(f(X^+)+f(X^-))\mathbb{I}_{\{f(X^+) > f(X^-) \}}] - \mathbb{E}[f(X^-)].    
\end{eqnarray*}

Given the initial assumptions , we can expand the first term $\mathbb{E}[f(X^+)\mathbb{I}_{\{f(X^+) > f(X^-) \}}]$ as follows:
\begin{eqnarray*}
\mathbb{E}[f(X^+)\mathbb{I}_{\{f(X^+) > f(X^-) \}}] = \\
\iint f(X^+)\mathbb{I}_{\{f(X^+) > f(X^-) \}} dP(X^+)dP(X^-)
\end{eqnarray*}

As $X^+$ and $X^-$ follow the same distribution, and $f$ is a measurable function, we use the \textit{Fubini} theorem to intervert the integrals as follows:

\begin{eqnarray*}
\mathbb{E}[f(X^+)\mathbb{I}_{\{f(X^+) > f(X^-) \}}] = \\
\iint f(X^-)\mathbb{I}_{\{f(X^-) \geq f(X^+) \}} dP(X^-)dP(X^+).
\end{eqnarray*}
 
Now replacing this in the original equation gives us:

\begin{eqnarray*}
\mathbb{E}[Z] &=& \iint f(X^-)\mathbb{I}_{\{f(X^-) \geq f(X^+) \}} dP(X^-)dP(X^+) \\ 
&& +\iint f(X^-)\mathbb{I}_{\{f(X^+) > f(X^-) \}} dP(X^-)dP(X^+)\\
&& -\mathbb{E}[f(X^-)]\\
&=& \mathbb{E}[(f(X^-)\mathbb{I}_{\{f(X^-) \geq f(X^+)\}}]\\
&&+\mathbb{E}[(f(X^-)\mathbb{I}_{\{f(X^+) > f(X^-) \}}] - \mathbb{E}[f(X^-)] \\
&=& 0
\end{eqnarray*}

This completes our proof that $Z$ is a centered random variable.
\end{proof}

Notice that setting the function $m = l\circ g\circ h$ we can write $l\circ g = f$ and $X = h(W)$. As we defined previously $h$ to be the weighted degree function of a node : 
 $$ h(W^+_j) = \sum_{i} W_{ij} \mathbb{I}_{\{W_{ij}>0\}} $$
 $$ h(W^-_j) = \sum_{i} |W_{ij}| \mathbb{I}_{\{W_{ij}\leq0\}}$$
which ensures that $h(W^+)$ and $h(W^-)$ follow the same distribution by linear combination of absolute value of the same normal distribution. 
Replacing these functions in the previous proposition, i.e. $f = l\circ g$, $X^+ =h(W^+) $ and $X^-=h(W^-)$ proves that our initialization method has mean $0$.
This proof allows us to justify the use of the Kaiming He  variance in our initialization method as it was proven to be the optimal one.

\section{Experiments}

We will now evaluate our proposed weight initialization  method Hcore-Init in image classification task using three standard datasets, CIFAR-$10$, CIFAR-$100$, and MNIST. We compare Hcore-Init to the results of Kaiming He initialization scheme. It is important to stress that we do not experiment on state-of-the-art architectures for each dataset. We want to show, as our method can be used separately on different architecture blocks, i.e. only at the convolutional layers, or only at the FCNN part, or both. We observe that it outperforms standard initialization methods, regardless of the block of the architecture that is initialized. Hence in this section, we  evaluate image classification accuracy on the aforementioned datasets with simple CNN architectures presented in this section. 

\subsection{Dataset specifications.}
The CIFAR-$10$ and CIFAR-$100$ datasets are labeled subsets of the $80$ million tiny images dataset collected by Alex Krizhevsky, Vinod Nair, and Geoffrey Hinton \cite{krizhevsky2009learning}.
\begin{itemize}
    \item The CIFAR-$10$ dataset consists of $60000$ $32\times32$ colour images in $10$ classes, with $6000$ images per class. There are $50000$ training images and $10000$ test images.
    \item The CIFAR-$100$ is just like the CIFAR-$10$, except it has $100$ classes containing $600$ images each. There are $500$ training images and $100$ testing images per class.
\end{itemize}
We also test our model on the MNIST database of handwritten digits, providing a training set of $60000$ examples, and a test set of $10 000$ examples. The digits have been size-normalized and centered in a fixed-size image \cite{lecun-mnisthandwrittendigit-2010}. 

The dataset is divided into five training batches and one test batch, each with $10000$ images. The test batch contains exactly $1000$ randomly-selected images from each class. The training batches contain the remaining images in random order, but some training batches may contain more images from one class than another. Between them, the training batches contain exactly $5000$ images from each class.

\subsection{Model setup and baseline.}
Next, we present the models that were trained and evaluated for the image classification task. We note that for every case, we compare two scenarios: 
\begin{enumerate}
    \item Initialization of the model with Kaiming He initialization \cite{he2015delving}, training on the train set for $150$ epochs and evaluation on the test set.
    \item Pretraining of the model (using Kaiming He initialization) for $N$ epochs, re-initialization of the model with Hcore-Init, training on the train set for the rest $150-N$ epochs and evaluation on the test set. $N$ has been set as a hyper-parameter.
\end{enumerate}

For the CIFAR-10 and CIFAR-100 datasets, we applied $2$ convolutional layers with sizes $3\times 6 \times 5$ and $6\times15\times 5$ respectively, where $5$ is the kernel size and the stride was set to $1$. Moreover, after each convolutional layer, we applied two $2\times 2$ max-pooling operators and finally three fully connected layers with corresponding sizes $400 \times 120$, $120 \times 84$, $84 \times$  $\# \text{classes}$, where  $ \# \text{classes} = 10 \text{ and } 100$ respectively for the two datasets. Furthermore, we used \textit{ReLU} as activation function among the linear layers and \textit{tanh} for the convolution layers.

For the MNIST dataset, we applied again $2$ convolutional layers of size $1\times 10 \times 5$ and $10\times 20 \times 5$, where again the filter size was set to $5$ and the stride was set to $1$. As in the other datasets, we employed two $2\times 2$ max-pooling operators and we performed dropout \cite{dropout} on the output of the $2^{nd}$ convolutional layer with probability $p=0.5$. Finally, we applied $2$ fully connected layers of size $320 \times 50$ and $50 \times 10$ and \textit{ReLU} as an activation function throughout the layers. 

In all cases, we employed stochastic gradient descent \cite{sgd}  with momentum set to $0.9$ and learning rate set to $0.001$. As we mentioned before, we chose $2$ rather simple models, as we intend to highlight the contribution of Hcore-Init in comparison to its competitor and not to achieve state-of-the-art results for the given datasets, which are exhaustively examined.

\subsection{Settings of the weight initialization.}
Next, we present the contribution of Hcore-Init  to the performance of the neural network architecture with respect to its application on different types of layers. Specifically, we applied the configurations of the initialization methods (a) exclusively on the set of the linear layers (b) exclusively on the set of the convolutional layers (c) on the combined set of linear and convolutional layers of the model. 

\begin{figure}[h!]
    \centering
    \includegraphics[width=9.5cm, height=4.5cm]{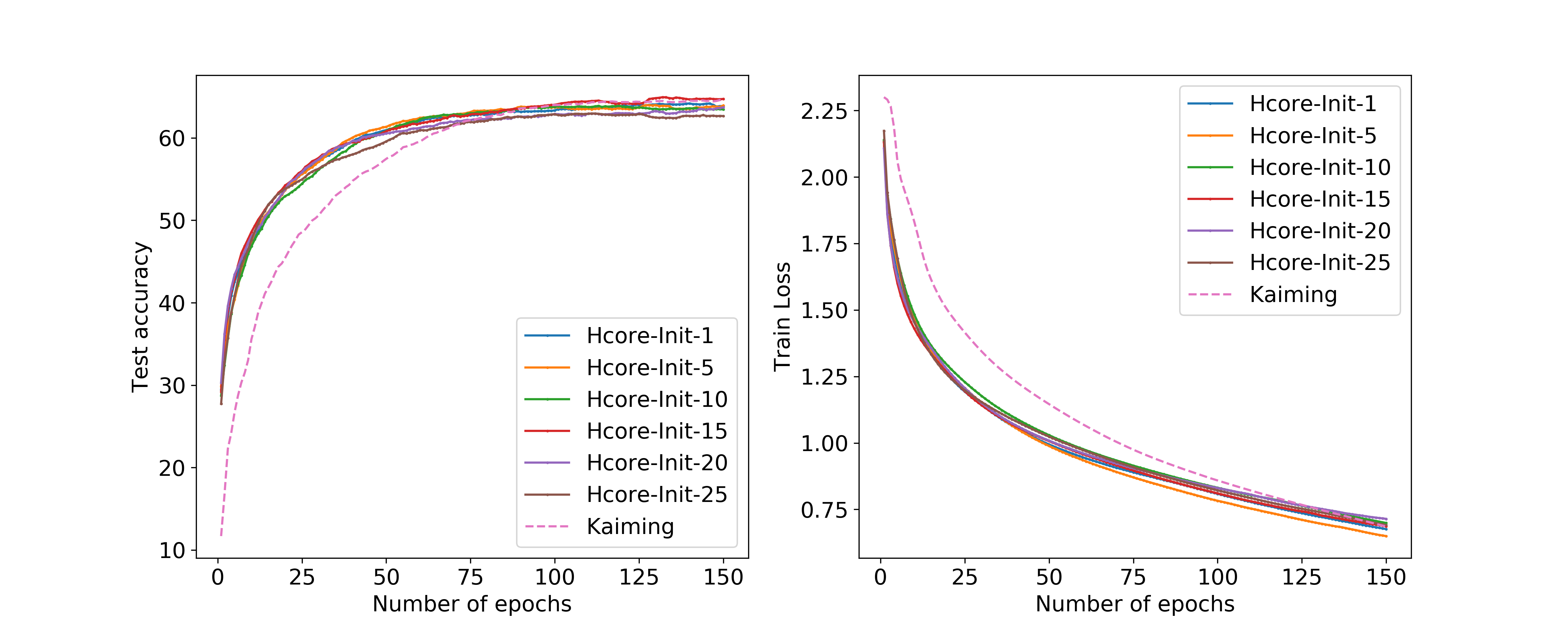}
    \caption{Test accuracy (left) and train loss (right) on CIFAR-$10$ for the combined application of the initialization on the linear and the convolutional layers. For the curves Hcore-init-$x$, $x$ stands for the number of pretraining epochs.}
    \label{fig:hcore_full}
\end{figure}

On Figure \ref{fig:hcore_full}, we observe that for $15$ pretraining epochs, the model initialized with Hcore-Init outperforms the model initialized with Kaiming He initialization. It is, also, noteworthy that the loss convergences faster when applying Hcore-Init. This highlights empirically our initial motivation of encouraging the ``important" weights by using the graph information from the model architecture.

\begin{figure}[h!]
    \centering
    \includegraphics[width=9.5cm, height=4.5cm]{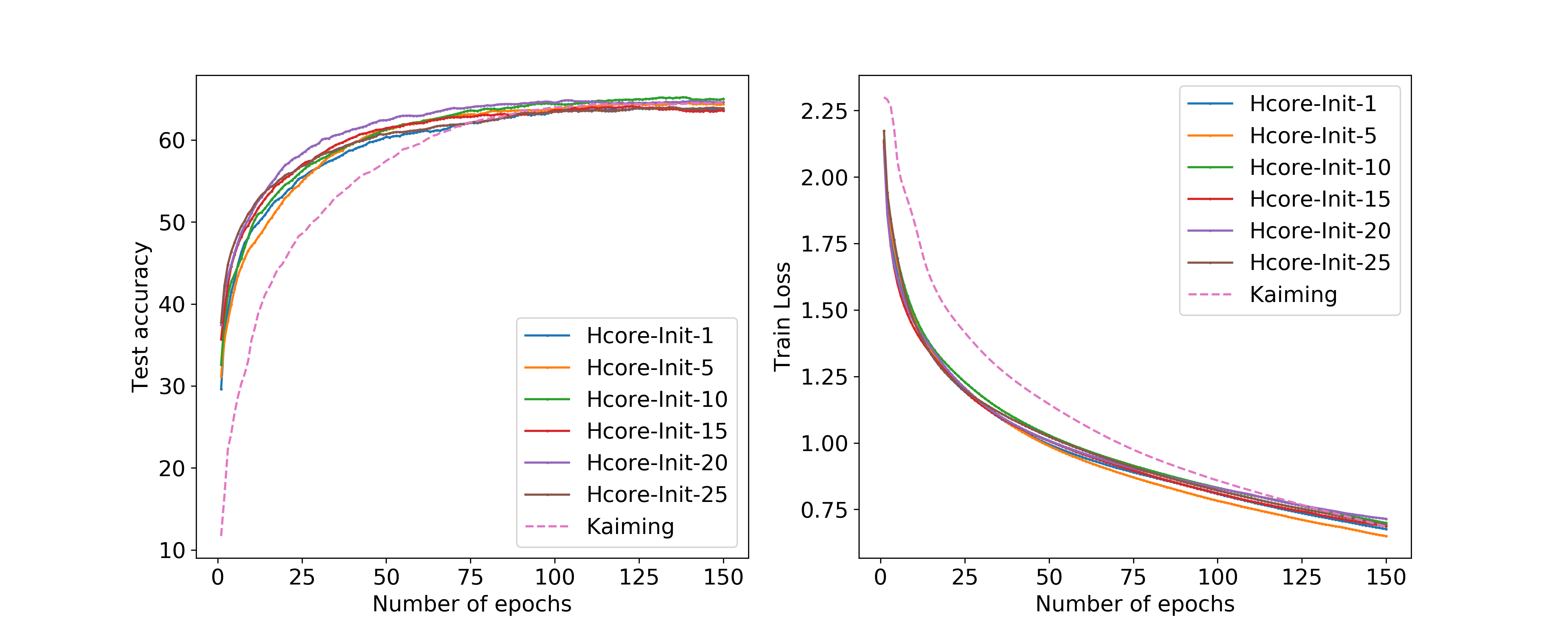}
    \caption{Test accuracy and train loss on CIFAR-$10$ for the initialization applied \textbf{only} on the linear layers.}
    \label{fig:hcore_lin}
\end{figure}

\begin{figure}[h!]
    \centering
    \includegraphics[width=9.5cm, height=4.5cm]{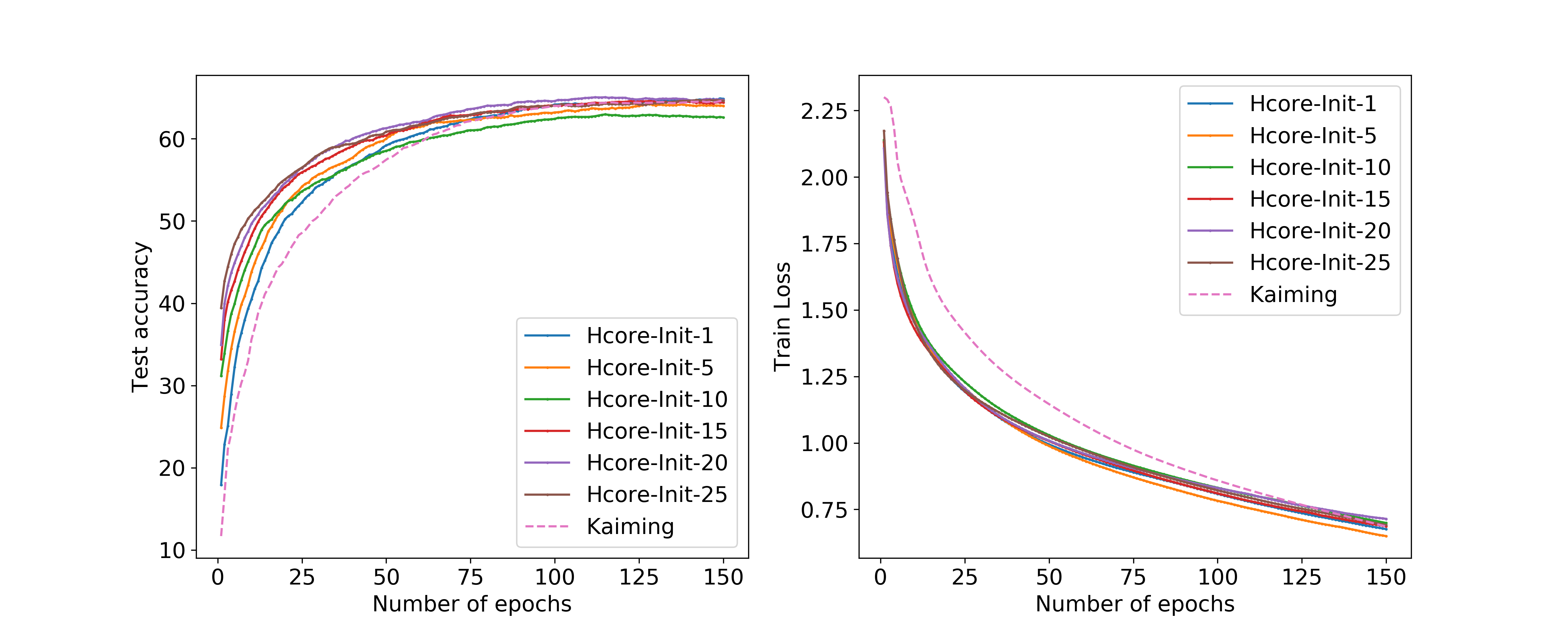}
    \caption{Test accuracy and train loss on CIFAR-$10$ for the initialization applied \textbf{only} on the convolutional layers.}
    \label{fig:hcore_cnn}
\end{figure}

On Figures \ref{fig:hcore_lin} and \ref{fig:hcore_cnn}, we can notice the contribution again of Hcore-Init in the performance of the network, when the former is applied on the fully connected and convolutional layers respectively. We can see that in both cases, Hcore-Init with different numbers of pretraining epochs ($10$ and $20$ correspondly) achieves better accuracy results in comparison to Kaiming He.  

\begin{table}[h!]
\centering
\caption{\textup{Top Accuracy results over initializing the full model, only the CNN and only the FCNN for CIFAR-$10$, CIFAR-$100$, and MNIST. \textbf{Hcore-Init*} represent the top performance over all the pretraining epochs configurations up to $25$}}

\begin{tabular}{|l|l|l|l|}

\hline
\multicolumn{1}{|c|}{}                 & \multicolumn{1}{c|}{\textbf{CIFAR-10}} & \multicolumn{1}{c|}{\textbf{CIFAR-100}} & \multicolumn{1}{c|}{\textbf{MNIST}} \\ \hline\hline
\multicolumn{1}{|c|}{\textbf{Kaiming He}} & \multicolumn{1}{c|}{$64.62$}                  & \multicolumn{1}{c|}{$32.56$}     & \multicolumn{1}{c|}{$98,71$}    \\ \hline

\multicolumn{1}{|c|}{\textbf{Hcore-Init*}} & \multicolumn{1}{c|}{$\mathbf{65.22}$}                  & \multicolumn{1}{c|}{$\mathbf{33.48}$}     & \multicolumn{1}{c|}{$\mathbf{98.91}$}    \\ \hline \hline
\multicolumn{1}{|c|}{\textbf{Hcore-Init-1}} & \multicolumn{1}{c|}{$64.91$}                  & \multicolumn{1}{c|}{$32.87$}                   & \multicolumn{1}{c|}{$98.59$}   \\ \hline
\multicolumn{1}{|c|}{\textbf{Hcore-Init-5}}    & \multicolumn{1}{c|}{$64.41$}   &    \multicolumn{1}{c|}{$32.96$}   &         \multicolumn{1}{c|}{$98.70$}    \\ \hline
\multicolumn{1}{|c|}{\textbf{Hcore-Init-10}}   &  \multicolumn{1}{c|}{$\mathbf{65.22}$}  &    
\multicolumn{1}{c|}{$33.41$} &     \multicolumn{1}{c|}{$98.81$}        \\ \hline
\multicolumn{1}{|c|}{\textbf{Hcore-Init-15}}   & \multicolumn{1}{c|}{$64.94$} &    
\multicolumn{1}{c|}{$33.45$} &     \multicolumn{1}{c|}{$98.64$}        \\ \hline
\multicolumn{1}{|c|}{\textbf{Hcore-Init-20}}   & \multicolumn{1}{c|}{$65.05$}   &    \multicolumn{1}{c|}{$33.39$}  &     \multicolumn{1}{c|}{$98.87$}       \\ \hline
\multicolumn{1}{|c|}{\textbf{Hcore-Init-25}}   & \multicolumn{1}{c|}{$64.72$}  &       \multicolumn{1}{c|}{$\mathbf{33.48}$}   &     \multicolumn{1}{c|}{$\mathbf{98.91}$}        \\ \hline

\end{tabular}
\label{table:resultsfull}

\end{table}
Finally, we report the results of the experiments conducted on the $3$ datasets in \ref{table:resultsfull}. Those results correspond to an ablation study over the different number of pretraining epochs as well as the different initialization scenarios, i.e. initializing only on the linear layers, convolutional layers, and the whole architecture. We kept for each mentioned scenario the best performance, and as it is evident \textbf{Hcore-Init*} achieves the best overall accuracy. It is important to stress that we do not necessarily need a long pretraining phase to achieve the best results, in fact, only $10$ epochs is usually more than enough to outperform in a significant way the Kaiming He initialization. We remind that this pretraining corresponds to less than $10\%$ of the total training which is proportional, in terms of computation time, to $10 \%$ of the time to train the model. Furthermore it is interesting to notice that in the early stages of pretraining we are more likely to lose some accuracy as the gradient direction in this stage of the training might be wrong. This justifies as well the consistency of our method.

\section{Conclusion} 

In this paper, we propose \textbf{Hcore-Init}, a novel  initialization method applicable on the most common blocks of neural network architectures, i.e. convolutional and linear layers. This method capitalizes on a graph representation of the neural network and more specifically on the densest parts of it found by the hypergraph degeneracy methods we define, providing thus a neuron ranking for the bipartite architecture of the neural network layers. Our method, learning with a small pretraining of the neural network, outperforms the  state of the art Kaiming He initialization, under the condition that the initialization distribution has zero expectancy. This work is intended to be used as a framework to initialize specific blocks in more complex architectures that might bear more information and are more valuable for the task at hand.

\section*{Acknowledgments}

We gratefully acknowledge the support of NVIDIA Corporation with the donation of the Titan Xp used for this research.






\bibliographystyle{IEEEtran}
\bibliography{biblio}

\begin{thebibliography}{10}

\bibitem{al2017identification}
Mohammed~Ali Al-garadi, Kasturi~Dewi Varathan, and Sri~Devi Ravana.
\newblock Identification of influential spreaders in online social networks
  using interaction weighted k-core decomposition method.
\newblock {\em Physica A: Statistical Mechanics and its Applications},
  468:278--288, 2017.

\bibitem{batagelj2003m}
Vladimir Batagelj and Matjaz Zaversnik.
\newblock An o (m) algorithm for cores decomposition of networks.
\newblock {\em arXiv preprint cs/0310049}, 2003.

\bibitem{giatsidis2013d}
Christos Giatsidis, Dimitrios~M Thilikos, and Michalis Vazirgiannis.
\newblock D-cores: measuring collaboration of directed graphs based on
  degeneracy.
\newblock {\em Knowledge and information systems}, 35(2):311--343, 2013.

\bibitem{Glorot10understandingthe}
Xavier Glorot and Yoshua Bengio.
\newblock Understanding the difficulty of training deep feedforward neural
  networks.
\newblock In {\em In Proceedings of the International Conference on Artificial
  Intelligence and Statistics (AISTATS’10). Society for Artificial
  Intelligence and Statistics}, 2010.

\bibitem{he2015delving}
Kaiming He, Xiangyu Zhang, Shaoqing Ren, and Jian Sun.
\newblock Delving deep into rectifiers: Surpassing human-level performance on
  imagenet classification, 2015.

\bibitem{sgd}
J.~Kiefer and J.~Wolfowitz.
\newblock Stochastic estimation of the maximum of a regression function.
\newblock {\em Ann. Math. Statist.}, 23(3):462--466, 09 1952.

\bibitem{krizhevsky2009learning}
Alex Krizhevsky et~al.
\newblock Learning multiple layers of features from tiny images.
\newblock 2009.

\bibitem{krizhevsky2012imagenet}
Alex Krizhevsky, Ilya Sutskever, and Geoffrey~E Hinton.
\newblock Imagenet classification with deep convolutional neural networks.
\newblock In {\em Advances in neural information processing systems}, pages
  1097--1105, 2012.

\bibitem{lecun-mnisthandwrittendigit-2010}
Yann LeCun and Corinna Cortes.
\newblock {MNIST} handwritten digit database.
\newblock 2010.

\bibitem{mishkin2015all}
Dmytro Mishkin and Jiri Matas.
\newblock All you need is a good init.
\newblock {\em arXiv preprint arXiv:1511.06422}, 2015.

\bibitem{morris2019weisfeiler}
Christopher Morris, Martin Ritzert, Matthias Fey, William~L Hamilton, Jan~Eric
  Lenssen, Gaurav Rattan, and Martin Grohe.
\newblock Weisfeiler and leman go neural: Higher-order graph neural networks.
\newblock In {\em Proceedings of the AAAI Conference on Artificial
  Intelligence}, volume~33, pages 4602--4609, 2019.

\bibitem{nikolentzos2018}
Giannis Nikolentzos, Polykarpos Meladianos, Stratis Limnios, and Michalis
  Vazirgiannis.
\newblock A degeneracy framework for graph similarity.
\newblock In {\em Proceedings of the Twenty-Seventh International Joint
  Conference on Artificial Intelligence, {IJCAI-18}}, pages 2595--2601.
  International Joint Conferences on Artificial Intelligence Organization, 7
  2018.

\bibitem{pisanski2000bridges}
Tomaz Pisanski and Milan Randic.
\newblock Bridges between geometry and graph theory.
\newblock {\em MAA NOTES}, pages 174--194, 2000.

\bibitem{rossi2015spread}
Maria-Evgenia~G Rossi, Fragkiskos~D Malliaros, and Michalis Vazirgiannis.
\newblock Spread it good, spread it fast: Identification of influential nodes
  in social networks.
\newblock In {\em Proceedings of the 24th International Conference on World
  Wide Web}, pages 101--102, 2015.

\bibitem{seidman1983network}
Stephen~B Seidman.
\newblock Network structure and minimum degree.
\newblock {\em Social networks}, 5(3):269--287, 1983.

\bibitem{skianis2019rep}
Konstantinos Skianis, Giannis Nikolentzos, Stratis Limnios, and Michalis
  Vazirgiannis.
\newblock Rep the set: Neural networks for learning set representations.
\newblock {\em arXiv preprint arXiv:1904.01962}, 2019.

\bibitem{dropout}
Nitish Srivastava, Geoffrey Hinton, Alex Krizhevsky, Ilya Sutskever, and Ruslan
  Salakhutdinov.
\newblock Dropout: A simple way to prevent neural networks from overfitting.
\newblock {\em J. Mach. Learn. Res.}, 15(1):1929–1958, January 2014.

\end{thebibliography}
%


\end{document}